\documentclass[10pt,twocolumn,letterpaper]{article}
\pdfoutput=1 

\usepackage{cvpr}
\usepackage{times}
\usepackage{epsfig}
\usepackage{graphicx}
\usepackage{amsmath}
\usepackage{amssymb}
\usepackage{textcomp}
\usepackage{algorithmic}
\usepackage{algorithm}
\usepackage{float}
\usepackage{pifont}
\usepackage{enumitem}
\usepackage[font=small,skip=0pt]{caption}
\usepackage{subcaption}
\usepackage{nicefrac}
\usepackage{microtype}

\usepackage{color}

\usepackage{amssymb}
\usepackage{amsfonts}
\usepackage{mathrsfs}
\usepackage{xspace}
\usepackage{bm}
\usepackage{upgreek}

\newcommand{\safemath}[2]{\newcommand{#1}{\ensuremath{#2}\xspace}}

\safemath{\bma}{\mathbf{a}}
\safemath{\bmb}{\mathbf{b}}
\safemath{\bmc}{\mathbf{c}}
\safemath{\bmd}{\mathbf{d}}
\safemath{\bme}{\mathbf{e}}
\safemath{\bmf}{\mathbf{f}}
\safemath{\bmg}{\mathbf{g}}
\safemath{\bmh}{\mathbf{h}}
\safemath{\bmi}{\mathbf{i}}
\safemath{\bmj}{\mathbf{j}}
\safemath{\bmk}{\mathbf{k}}
\safemath{\bml}{\mathbf{l}}
\safemath{\bmm}{\mathbf{m}}
\safemath{\bmn}{\mathbf{n}}
\safemath{\bmo}{\mathbf{o}}
\safemath{\bmp}{\mathbf{p}}
\safemath{\bmq}{\mathbf{q}}
\safemath{\bmr}{\mathbf{r}}
\safemath{\bms}{\mathbf{s}}
\safemath{\bmt}{\mathbf{t}}
\safemath{\bmu}{\mathbf{u}}
\safemath{\bmv}{\mathbf{v}}
\safemath{\bmw}{\mathbf{w}}
\safemath{\bmx}{\mathbf{x}}
\safemath{\bmy}{\mathbf{y}}
\safemath{\bmz}{\mathbf{z}}
\safemath{\bmzero}{\mathbf{0}}
\safemath{\bmone}{\mathbf{1}}

\bmdefine{\biad}{a}
\bmdefine{\bibd}{b}
\bmdefine{\bicd}{c}
\bmdefine{\bidd}{d}
\bmdefine{\bied}{e}
\bmdefine{\bifd}{f}
\bmdefine{\bigd}{g}
\bmdefine{\bihd}{h}
\bmdefine{\biid}{i}
\bmdefine{\bijd}{j}
\bmdefine{\bikd}{k}
\bmdefine{\bild}{l}
\bmdefine{\bimd}{m}
\bmdefine{\bind}{n}
\bmdefine{\biod}{o}
\bmdefine{\bipd}{p}
\bmdefine{\biqd}{q}
\bmdefine{\bird}{r}
\bmdefine{\bisd}{s}
\bmdefine{\bitd}{t}
\bmdefine{\biud}{u}
\bmdefine{\bivd}{v}
\bmdefine{\biwd}{w}
\bmdefine{\bixd}{x}
\bmdefine{\biyd}{y}
\bmdefine{\bizd}{z}

\bmdefine{\bixid}{\xi}
\bmdefine{\bilambdad}{\lambda}
\bmdefine{\bimud}{\mu}
\bmdefine{\bithetad}{\theta}
\bmdefine{\biphid}{\phi}
\bmdefine{\bideltad}{\delta}

\safemath{\bmia}{\biad}
\safemath{\bmib}{\bibd}
\safemath{\bmic}{\bicd}
\safemath{\bmid}{\bidd}
\safemath{\bmie}{\bied}
\safemath{\bmif}{\bifd}
\safemath{\bmig}{\bigd}
\safemath{\bmih}{\bihd}
\safemath{\bmii}{\biid}
\safemath{\bmij}{\bijd}
\safemath{\bmik}{\bikd}
\safemath{\bmil}{\bild}
\safemath{\bmim}{\bimd}
\safemath{\bmin}{\bind}
\safemath{\bmio}{\biod}
\safemath{\bmip}{\bipd}
\safemath{\bmiq}{\biqd}
\safemath{\bmir}{\bird}
\safemath{\bmis}{\bisd}
\safemath{\bmit}{\bitd}
\safemath{\bmiu}{\biud}
\safemath{\bmiv}{\bivd}
\safemath{\bmiw}{\biwd}
\safemath{\bmix}{\bixd}
\safemath{\bmiy}{\biyd}
\safemath{\bmiz}{\bizd}

\safemath{\bmxi}{\bixid}
\safemath{\bmlambda}{\bilambdad}
\safemath{\bmmu}{\bimud}
\safemath{\bmtheta}{\bithetad}
\safemath{\bmphi}{\biphid}
\safemath{\bmdelta}{\bideltad}

\safemath{\bA}{\mathbf{A}}
\safemath{\bB}{\mathbf{B}}
\safemath{\bC}{\mathbf{C}}
\safemath{\bD}{\mathbf{D}}
\safemath{\bE}{\mathbf{E}}
\safemath{\bF}{\mathbf{F}}
\safemath{\bG}{\mathbf{G}}
\safemath{\bH}{\mathbf{H}}
\safemath{\bI}{\mathbf{I}}
\safemath{\bJ}{\mathbf{J}}
\safemath{\bK}{\mathbf{K}}
\safemath{\bL}{\mathbf{L}}
\safemath{\bM}{\mathbf{M}}
\safemath{\bN}{\mathbf{N}}
\safemath{\bO}{\mathbf{O}}
\safemath{\bP}{\mathbf{P}}
\safemath{\bQ}{\mathbf{Q}}
\safemath{\bR}{\mathbf{R}}
\safemath{\bS}{\mathbf{S}}
\safemath{\bT}{\mathbf{T}}
\safemath{\bU}{\mathbf{U}}
\safemath{\bV}{\mathbf{V}}
\safemath{\bW}{\mathbf{W}}
\safemath{\bX}{\mathbf{X}}
\safemath{\bY}{\mathbf{Y}}
\safemath{\bZ}{\mathbf{Z}}

\safemath{\bZero}{\mathbf{0}}
\safemath{\bOne}{\mathbf{1}}
\safemath{\bDelta}{\mathbf{\Delta}}
\safemath{\bLambda}{\mathbf{\UpLambda}}
\safemath{\bPhi}{\mathbf{\Phi}}
\safemath{\bSigma}{\mathbf{\Upsigma}}
\safemath{\bOmega}{\mathbf{\Upomega}}
\safemath{\bTheta}{\mathbf{\Uptheta}}

\bmdefine{\biAd}{A}
\bmdefine{\biBd}{B}
\bmdefine{\biCd}{C}
\bmdefine{\biDd}{D}
\bmdefine{\biEd}{E}
\bmdefine{\biFd}{F}
\bmdefine{\biGd}{G}
\bmdefine{\biHd}{H}
\bmdefine{\biId}{I}
\bmdefine{\biJd}{J}
\bmdefine{\biKd}{K}
\bmdefine{\biLd}{L}
\bmdefine{\biMd}{M}
\bmdefine{\biOd}{N}
\bmdefine{\biPd}{O}
\bmdefine{\biQd}{P}
\bmdefine{\biRd}{R}
\bmdefine{\biSd}{S}
\bmdefine{\biTd}{T}
\bmdefine{\biUd}{U}
\bmdefine{\biVd}{V}
\bmdefine{\biWd}{W}
\bmdefine{\biXd}{X}
\bmdefine{\biYd}{Y}
\bmdefine{\biZd}{Z}

\bmdefine{\biDelta}{\Delta}
\bmdefine{\biLambda}{\Lambda}
\bmdefine{\biPhi}{\Phi}
\bmdefine{\biSigma}{\Sigma}
\bmdefine{\biOmega}{\Omega}
\bmdefine{\biTheta}{\Theta}

\safemath{\bimA}{\biAd}
\safemath{\bimB}{\biBd}
\safemath{\bimC}{\biCd}
\safemath{\bimD}{\biDd}
\safemath{\bimE}{\biEd}
\safemath{\bimF}{\biFd}
\safemath{\bimG}{\biGd}
\safemath{\bimH}{\biHd}
\safemath{\bimI}{\biId}
\safemath{\bimJ}{\biJd}
\safemath{\bimK}{\biKd}
\safemath{\bimL}{\biLd}
\safemath{\bimM}{\biMd}
\safemath{\bimN}{\biNd}
\safemath{\bimO}{\biOd}
\safemath{\bimP}{\biPd}
\safemath{\bimQ}{\biQd}
\safemath{\bimR}{\biRd}
\safemath{\bimS}{\biSd}
\safemath{\bimT}{\biTd}
\safemath{\bimU}{\biUd}
\safemath{\bimV}{\biVd}
\safemath{\bimW}{\biWd}
\safemath{\bimX}{\biXd}
\safemath{\bimY}{\biYd}
\safemath{\bimZ}{\biZd}

\safemath{\bimDelta}{\biDelta}
\safemath{\bimLambda}{\biLambda}
\safemath{\bimPhi}{\biPhi}
\safemath{\bimSigma}{\biSigma}
\safemath{\bimOmega}{\biOmega}
\safemath{\bimTheta}{\biTheta}

\safemath{\setA}{\mathcal{A}}
\safemath{\setB}{\mathcal{B}}
\safemath{\setC}{\mathcal{C}}
\safemath{\setD}{\mathcal{D}}
\safemath{\setE}{\mathcal{E}}
\safemath{\setF}{\mathcal{F}}
\safemath{\setG}{\mathcal{G}}
\safemath{\setH}{\mathcal{H}}
\safemath{\setI}{\mathcal{I}}
\safemath{\setJ}{\mathcal{J}}
\safemath{\setK}{\mathcal{K}}
\safemath{\setL}{\mathcal{L}}
\safemath{\setM}{\mathcal{M}}
\safemath{\setN}{\mathcal{N}}
\safemath{\setO}{\mathcal{O}}
\safemath{\setP}{\mathcal{P}}
\safemath{\setQ}{\mathcal{Q}}
\safemath{\setR}{\mathcal{R}}
\safemath{\setS}{\mathcal{S}}
\safemath{\setT}{\mathcal{T}}
\safemath{\setU}{\mathcal{U}}
\safemath{\setV}{\mathcal{V}}
\safemath{\setW}{\mathcal{W}}
\safemath{\setX}{\mathcal{X}}
\safemath{\setY}{\mathcal{Y}}
\safemath{\setZ}{\mathcal{Z}}
\safemath{\emptySet}{\varnothing}

\safemath{\colA}{\mathscr{A}}
\safemath{\colB}{\mathscr{B}}
\safemath{\colC}{\mathscr{C}}
\safemath{\colD}{\mathscr{D}}
\safemath{\colE}{\mathscr{E}}
\safemath{\colF}{\mathscr{F}}
\safemath{\colG}{\mathscr{G}}
\safemath{\colH}{\mathscr{H}}
\safemath{\colI}{\mathscr{I}}
\safemath{\colJ}{\mathscr{J}}
\safemath{\colK}{\mathscr{K}}
\safemath{\colL}{\mathscr{L}}
\safemath{\colM}{\mathscr{M}}
\safemath{\colN}{\mathscr{N}}
\safemath{\colO}{\mathscr{O}}
\safemath{\colP}{\mathscr{P}}
\safemath{\colQ}{\mathscr{Q}}
\safemath{\colR}{\mathscr{R}}
\safemath{\colS}{\mathscr{S}}
\safemath{\colT}{\mathscr{T}}
\safemath{\colU}{\mathscr{U}}
\safemath{\colV}{\mathscr{V}}
\safemath{\colW}{\mathscr{W}}
\safemath{\colX}{\mathscr{X}}
\safemath{\colY}{\mathscr{Y}}
\safemath{\colZ}{\mathscr{Z}}

\safemath{\opA}{\mathbb{A}}
\safemath{\opB}{\mathbb{B}}
\safemath{\opC}{\mathbb{C}}
\safemath{\opD}{\mathbb{D}}
\safemath{\opE}{\mathbb{E}}
\safemath{\opF}{\mathbb{F}}
\safemath{\opG}{\mathbb{G}}
\safemath{\opH}{\mathbb{H}}
\safemath{\opI}{\mathbb{I}}
\safemath{\opJ}{\mathbb{J}}
\safemath{\opK}{\mathbb{K}}
\safemath{\opL}{\mathbb{L}}
\safemath{\opM}{\mathbb{M}}
\safemath{\opN}{\mathbb{N}}
\safemath{\opO}{\mathbb{O}}
\safemath{\opP}{\mathbb{P}}
\safemath{\opQ}{\mathbb{Q}}
\safemath{\opR}{\mathbb{R}}
\safemath{\opS}{\mathbb{S}}
\safemath{\opT}{\mathbb{T}}
\safemath{\opU}{\mathbb{U}}
\safemath{\opV}{\mathbb{V}}
\safemath{\opW}{\mathbb{W}}
\safemath{\opX}{\mathbb{X}}
\safemath{\opY}{\mathbb{Y}}
\safemath{\opZ}{\mathbb{Z}}
\safemath{\opZero}{\mathbb{O}}
\safemath{\identityop}{\opI}

\safemath{\veca}{\bma}
\safemath{\vecb}{\bmb}
\safemath{\vecc}{\bmc}
\safemath{\vecd}{\bmd}
\safemath{\vece}{\bme}
\safemath{\vecf}{\bmf}
\safemath{\vecg}{\bmg}
\safemath{\vech}{\bmh}
\safemath{\veci}{\bmi}
\safemath{\vecj}{\bmj}
\safemath{\veck}{\bmk}
\safemath{\vecl}{\bml}
\safemath{\vecm}{\bmm}
\safemath{\vecn}{\bmn}
\safemath{\veco}{\bmo}
\safemath{\vecp}{\bmp}
\safemath{\vecq}{\bmq}
\safemath{\vecr}{\bmr}
\safemath{\vecs}{\bms}
\safemath{\vect}{\bmt}
\safemath{\vecu}{\bmu}
\safemath{\vecv}{\bmv}
\safemath{\vecw}{\bmw}
\safemath{\vecx}{\bmx}
\safemath{\vecy}{\bmy}
\safemath{\vecz}{\bmz}

\safemath{\veczero}{\bmzero}
\safemath{\vecone}{\bmone}
\safemath{\vecxi}{\bmxi}
\safemath{\veclambda}{\bmlambda}
\safemath{\vecmu}{\bmmu}
\safemath{\vectheta}{\bmtheta}
\safemath{\vecphi}{\bmphi}
\safemath{\vecdelta}{\bmdelta}

\safemath{\matA}{\bA}
\safemath{\matB}{\bB}
\safemath{\matC}{\bC}
\safemath{\matD}{\bD}
\safemath{\matE}{\bE}
\safemath{\matF}{\bF}
\safemath{\matG}{\bG}
\safemath{\matH}{\bH}
\safemath{\matI}{\bI}
\safemath{\matJ}{\bJ}
\safemath{\matK}{\bK}
\safemath{\matL}{\bL}
\safemath{\matM}{\bM}
\safemath{\matN}{\bN}
\safemath{\matO}{\bO}
\safemath{\matP}{\bP}
\safemath{\matQ}{\bQ}
\safemath{\matR}{\bR}
\safemath{\matS}{\bS}
\safemath{\matT}{\bT}
\safemath{\matU}{\bU}
\safemath{\matV}{\bV}
\safemath{\matW}{\bW}
\safemath{\matX}{\bX}
\safemath{\matY}{\bY}
\safemath{\matZ}{\bZ}
\safemath{\matzero}{\bmzero}

\safemath{\matDelta}{\bDelta}
\safemath{\matLambda}{\bLambda}
\safemath{\matPhi}{\bPhi}
\safemath{\matSigma}{\bSigma}
\safemath{\matOmega}{\bOmega}
\safemath{\matTheta}{\bTheta}

\safemath{\matidentity}{\matI}
\safemath{\matone}{\matO}

\safemath{\rnda}{A}
\safemath{\rndb}{B}
\safemath{\rndc}{C}
\safemath{\rndd}{D}
\safemath{\rnde}{E}
\safemath{\rndf}{F}
\safemath{\rndg}{G}
\safemath{\rndh}{H}
\safemath{\rndi}{I}
\safemath{\rndj}{J}
\safemath{\rndk}{K}
\safemath{\rndl}{L}
\safemath{\rndm}{M}
\safemath{\rndn}{N}
\safemath{\rndo}{O}
\safemath{\rndp}{P}
\safemath{\rndq}{Q}
\safemath{\rndr}{R}
\safemath{\rnds}{S}
\safemath{\rndt}{T}
\safemath{\rndu}{U}
\safemath{\rndv}{V}
\safemath{\rndw}{W}
\safemath{\rndx}{X}
\safemath{\rndy}{Y}
\safemath{\rndz}{Z}

\safemath{\rveca}{\bimA}
\safemath{\rvecb}{\bimB}
\safemath{\rvecc}{\bimC}
\safemath{\rvecd}{\bimD}
\safemath{\rvece}{\bimE}
\safemath{\rvecf}{\bimF}
\safemath{\rvecg}{\bimG}
\safemath{\rvech}{\bimH}
\safemath{\rveci}{\bimI}
\safemath{\rvecj}{\bimJ}
\safemath{\rveck}{\bimK}
\safemath{\rvecl}{\bimL}
\safemath{\rvecm}{\bimM}
\safemath{\rvecn}{\bimN}
\safemath{\rveco}{\bomO}
\safemath{\rvecp}{\bimP}
\safemath{\rvecq}{\bimQ}
\safemath{\rvecr}{\bimR}
\safemath{\rvecs}{\bimS}
\safemath{\rvect}{\bimT}
\safemath{\rvecu}{\bimU}
\safemath{\rvecv}{\bimV}
\safemath{\rvecw}{\bimW}
\safemath{\rvecx}{\bimX}
\safemath{\rvecy}{\bimY}
\safemath{\rvecz}{\bimZ}

\safemath{\rvecxi}{\bmxi}
\safemath{\rveclambda}{\bmlambda}
\safemath{\rvecmu}{\bmmu}
\safemath{\rvectheta}{\bmtheta}
\safemath{\rvecphi}{\bmphi}

\safemath{\rmatA}{\bimA}
\safemath{\rmatB}{\bimB}
\safemath{\rmatC}{\bimC}
\safemath{\rmatD}{\bimD}
\safemath{\rmatE}{\bimE}
\safemath{\rmatF}{\bimF}
\safemath{\rmatG}{\bimG}
\safemath{\rmatH}{\bimH}
\safemath{\rmatI}{\bimI}
\safemath{\rmatJ}{\bimJ}
\safemath{\rmatK}{\bimK}
\safemath{\rmatL}{\bimL}
\safemath{\rmatM}{\bimM}
\safemath{\rmatN}{\bimN}
\safemath{\rmatO}{\bimO}
\safemath{\rmatP}{\bimP}
\safemath{\rmatQ}{\bimQ}
\safemath{\rmatR}{\bimR}
\safemath{\rmatS}{\bimS}
\safemath{\rmatT}{\bimT}
\safemath{\rmatU}{\bimU}
\safemath{\rmatV}{\bimV}
\safemath{\rmatW}{\bimW}
\safemath{\rmatX}{\bimX}
\safemath{\rmatY}{\bimY}
\safemath{\rmatZ}{\bimZ}

\safemath{\rmatDelta}{\bimDelta}
\safemath{\rmatLambda}{\bimLambda}
\safemath{\rmatPhi}{\bimPhi}
\safemath{\rmatSigma}{\bimSigma}
\safemath{\rmatOmega}{\bimOmega}
\safemath{\rmatTheta}{\bimTheta}

\newcommand{\eqn}[2]{\begin{equation}\label{#1}#2\end{equation}}

\newcommand{\aln}[1]{\begin{align}#1\end{align}}

\newcommand{\matb}{\left( \begin{matrix*}[r] }
\newcommand{\mate}{\end{matrix*}\right)}

\newcommand{\half}{\frac{1}{2}}

\newcommand{\opt}{^\star}
\newcommand{\reals}{\mathbb{R}}

\newcommand{\ellone}{$\ell_1$}
\newcommand{\elltwo}{$\ell_2$}

\newcommand{\st}{\hbox{ \,\,subject to\,\, }}

\DeclareMathOperator{\supp}{supp}

\DeclareMathOperator*{\argmin}{arg\,min}

\DeclareMathOperator{\prox}{prox}

\usepackage[pagebackref=true,breaklinks=true,letterpaper=true,colorlinks,bookmarks=false]{hyperref}

\cvprfinalcopy % *** Uncomment this line for the final submission

\def\cvprPaperID{1599} % *** Enter the CVPR Paper ID here

\ifcvprfinal\fi

\usepackage{etoolbox}
\newcommand{\zerodisplayskips}{%
  \setlength{\abovedisplayskip}{5pt}
  \setlength{\belowdisplayskip}{5pt}
  \setlength{\abovedisplayshortskip}{5pt}
  \setlength{\belowdisplayshortskip}{5pt}}
\appto{\normalsize}{\zerodisplayskips}
\appto{\small}{\zerodisplayskips}
\appto{\footnotesize}{\zerodisplayskips}

\begin{document}
\setlength{\textfloatsep}{8pt}
\setlength{\dbltextfloatsep}{10pt plus 0pt minus 2pt}

%%%%%%%%% TITLE
%\title{Convex Regularization of Sparse Signals with Smooth  Support}
%\title{$l_{1,2}$-Magic: Estimating Sparse Signals with Smooth Support via Convex Programming}
\title{Estimating Sparse Signals with Smooth Support\\ via Convex Programming and Block Sparsity}

\author{Sohil Shah$^1$, Tom Goldstein$^1$, and  Christoph Studer$^2$ \\
$^1$University of Maryland, College Park;
$^2$Cornell University, Ithaca, NY\\
{\tt\small sohilas@umd.edu,\,tomg@cs.umd.edu,\,studer@cornell.edu}
% \and
}
\maketitle
\thispagestyle{empty}

%%%%%%%%% ABSTRACT
\begin{abstract}
Conventional algorithms for sparse signal recovery and sparse representation rely on \ellone{}-norm regularized variational methods. However, when applied to the reconstruction of {\em sparse  images}, i.e., images where only a few pixels are non-zero, simple \ellone{}-norm-based methods ignore potential correlations in the support between adjacent pixels. In a number of applications, one is interested in images that are not only sparse, but also have a support with smooth (or contiguous) boundaries.  Existing algorithms that take into account such a support structure   mostly rely on non-convex methods and---as a consequence---do not scale well to high-dimensional problems and/or do not converge to global optima. In this paper, we explore the use of new block \ellone{}-norm regularizers, which enforce image sparsity while simultaneously promoting smooth support structure.  By exploiting  the convexity of our regularizers, we develop new computationally-efficient recovery algorithms that guarantee global optimality.  We demonstrate the efficacy of our regularizers on a variety of imaging tasks including compressive image recovery, image restoration, and robust PCA.   
\end{abstract}

%%%%%%%%% BODY TEXT
\section{Introduction}

A large number of existing models used in sparse signal processing and machine learning rely on \ellone{}-norm regularization in order to recover sparse signals or to identify sparse features for classification tasks.  Sparse \ellone{}-norm  regularization is also prominently used in the image-processing and computer vision domain, where it is used for segmentation, tracking, and background subtraction tasks. %, as well as in biomedical image analysis. 
In computer vision and image processing, we are often interested in regions that are not only sparse, but also {\em spatially smooth}, i.e., regions with contiguous support structure. In such situations, it is desirable to have regularizers that promote the selection of large, contiguous regions rather than merely sparse (and potentially isolated) pixels.  In contrast, simple \ellone{}-norm regularization adopts an unstructured approach that induces sparsity wherein each variable is treated independently, disregarding correlation among neighboring variables. 
For example,  smooth support structure is relevant to compressive background subtraction \cite{
cevher2008compressive, cevher2009sparse} which detects contiguous regions of movement against a stationary background.

For imaging applications, \ellone{}-norm regularization may result in regions with spurious active (or isolated) pixels or non-smooth boundaries in the support set. This issue is addressed by the image-segmentation literature, where spatially correlated priors (such as total variation or normalized cuts) are used to enforce smooth support boundaries~\cite{boykov2001fast,felzenszwalb2004efficient,shi2000normalized,comaniciu2002mean}.   An important hallmark of existing image-segmentation methods is that they are able to enforce spatially contiguous support.  However, the concept of correlated support has yet to be ported to more complex reconstruction tasks, including (but not limited to) robust PCA and compressive background subtraction.
The development of such structured sparsity models has been an active research topic \cite{cevher2009sparse,baraniuk2010model,huang2011learning,jenatton2011structured,bach2011convex,jacob2009group}, with new models and applications still emerging \cite{jeni2014spatio,jenatton2009structured}. 
In this paper, we develop a class of convex priors based on overlapping block/group sparsity, which are able to enforce   sparsity of the support set and promote {\em spatial smoothness}.

\subsection{Relevant Previous Work}

Existing work on spatially-smooth support-set regularization can be divided into two main categories: (i) non-convex models that rely on graphs and trees, and (ii) convex models that rely on group-sparsity inducing norms. 
Cevher \etal~\cite{cevher2009sparse} promote sparsity using Markov random fields (MRFs) in combination with compressive-sensing signal recovery, which is referred to as lattice matching pursuit (LaMP).  LaMP recovers structured sparse signals using fewer noisy measurements than methods that ignore spatially correlated support sets. Baraniuk \etal \cite{baraniuk2010model} prove theoretical guarantees on robust recovery of structured sparse signals using a non-convex algorithm; their approach has been validated using wavelet-tree-based hierarchical group structure, as well as signals with non-overlapping blocks in the support set. Huang \etal \cite{huang2011learning} developed a theory of greedy approximation methods  for general non-convex structured sparse models.
All these methods, however, are limited in that they are either non-convex, computationally expensive, or do not allow for overlapping (or not aligned) group structure.
 Jenatton \etal \cite{jenatton2011structured} showed the possibility of coming up with a problem-specific optimal group-sparsity-inducing norm using prior knowledge of the underlying structure. While they consider a convex relaxation of the structured sparsity problem, it remains unclear  how their proposed active-set algorithm for least squares regression can be generalized to a broader range of applications. 

\subsection{Contributions}
Our work is inspired by the $\ell_1/\ell_2$-norm spatial coherence priors used in \cite{jenatton2011structured}, as well as group sparsity priors used in statistics (e.g., group lasso) \cite{leigsnering2014multipath,yuan2006model}.  Our main contributions can be summarized as follows: (i) 
We propose new regularizers for imaging and computer vision applications including compressive image recovery, sparse \& low rank decomposition, and a block-sparse generalization of total variation. 
(ii) We develop computationally efficient global minimization algorithms that are suitable for overlapping pixel-cliques.  
Existing methods for group sparsity use the alternating direction method of multipliers (ADMM), and have excessive memory requirements for large clique sizes.  We therefore discuss a new approach using fast convolution algorithms to perform gradient descent with low memory requirements and a complexity that is independent of the clique size.  
(iii) We propose the use of our regularizers within greedy pursuit methods for compressive reconstruction.
%
%We make three specific contributions in this paper:
%\begin{itemize}[leftmargin=*]
% \item Our paper proposes the ADMM approach to solve any {\em general} problem with structured sparsity regularization prior enforced on overlapping cliques using mixed sparsity $\ell_1/\ell_2$ norm.
% \item Furthermore, we also propose forward-backward splitting algorithm to efficiently solve problems with structures of larger clique size.
% \item Finally, we demonstrate that our methods can be used to suppress artifacts and enhance the quality of sparse recovery methods when applied to various imaging applications such as compressed sensing recovery, gaussian denoising and robust PCA.
%\end{itemize}
%
(iv) We demonstrate that our algorithms can be used to suppress artifacts and enhance the quality of sparse recovery methods when applied to a variety of imaging applications.
%, including compressive sensing image recovery, Gaussian denoising, and robust PCA.

%------------------------------------------------------------------------

\subsection{Notation} 
For any column vector $\mathbf{x} \in \mathbb{R}^n$, we define its $\ell_{\alpha}$-norm with $\alpha\geq1$ as $\|\mathbf{x}\|_{\alpha} = (\sum_{i=1}^n |x_i|^{\alpha})^{1/{\alpha}}$.
For $\mathbf{x} \in \mathbb{R}^n$, the vector $\mathbf{x}_c$ consists only of the entries associated to the index set $c$. The support set (i.e., the set of indices of non-zero entries) of a vector or vectorized image $\vecx$ is denoted by $\supp(\vecx).$ 
 For a matrix $\mathbf{A} \in \mathbb{R}^{M\times N}$ with rank $r = \min\{M,N\}$ and singular values $\sigma_i$, the nuclear norm is defined by $\|\bA\|_* = \sum_{i=1}^r \sigma_i.$  We use $\|\bA\|_1 = \sum_{ij} |A_{ij}|$ to denote the element-wise \ellone{}-norm for $\mathbf{A}.$

%------------------------------------------------------------------------

\begin{figure}[tb]
\centering
   \includegraphics[width=0.9\linewidth]{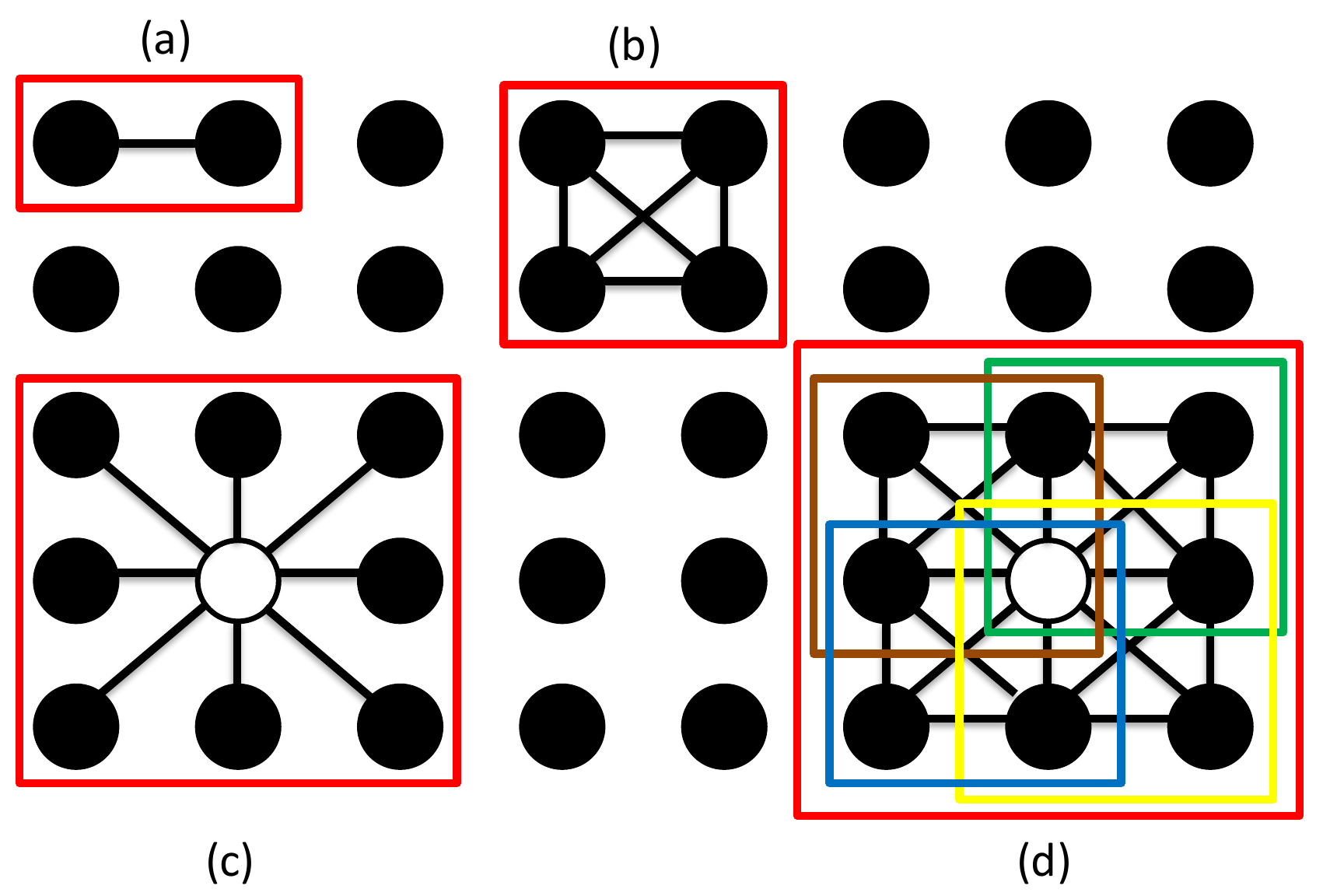} %height=2.5in
   \vspace{0.1cm}
   \caption{Illustration of cliques and overlapping cliques.}
\label{fig:cliques}
\end{figure}

\section{Problem Formulation}

Consider the measurement  model $\vecy = \matPhi \vecx_0 + \vecz_0$,
 where $\vecy \in \reals^M$ is the observed signal, $\mathbf{x}_0 \in \mathbb{R}^N$ is the original sparse signal we wish to recover, $\mathbf{z}_0 \in \mathbb{R}^M$ is a non-sparse component of the signal (comprising both the background image and potential noise), $\mathbf{\Phi} \in \mathbb{R}^{M\times N}$ is the linear operator that models the signal acquisition process. Based on this model, we study signal recovery by solving convex optimization problems of the following general form:
\begin{align}
 \{\hat\vecz,\hat\vecx\}= \argmin_{\mathbf{z} \in \mathbb{R}^{M}\!,\mathbf{x} \in \mathbb{R}^N} D(\vecx,\vecz\,|\,\vecy,\matPhi) + J(\vecx). \label{general}
\end{align}
Here, $D: \mathbb{R}^M\!\times  \mathbb{R}^M  \mathcal \to\, \mathbb{R}$ is a convex data-consistency term, and $J: \mathbb{R}^M \mathcal \to\, \mathbb{R}$ is a regularizer that enforces both sparsity and support smoothness on the vector $\vecx.$  The proposed regularizer is a hybrid $\ell_1/\ell_2$-norm penalty of the form
 \eqn{reg}{ \textstyle J(\vecx) = \sum_{c \in \mathcal{C}} \|\mathbf{x}_c\|_2,}
where $\mathcal{C}$ is a set of cliques over the graph $\mathcal{G}$ defined over the pixels of $\vecx.$ This regularizer \eqref{reg} is a natural generalization of the {\em group (or block) sparsity} model that has been explored in the literature for a variety of purposes including statistics and radar  \cite{jenatton2011structured,jacob2009group,jeni2014spatio,leigsnering2014multipath}. We focus on the case where the collection of sparse cliques consist of regularly-spaced groups of adjacent pixels.  
For example, consider two types of cliques shown in Figure \ref{fig:cliques}(a) and \ref{fig:cliques}(b).  Notice the (a) 2-clique and (b) 4-clique wherein all nodes are connected to each other. These cliques can be translated over the entire image graph to generate various overlapping clique geometries as shown in (c) and (d), respectively. In (c), eight overlapping cliques, each of size two, overlap at a central point.  In the image processing literature this is referred to as an 8-connected neighborhood~\cite{cheng2009subband}.  In contrast, Figure \ref{fig:cliques}(d) uses a higher-order connectivity model, which is obtained using four rectangular cliques of size four (each shown in a different color).  
  Overlapping group-sparsity models of the form depicted in Figure \ref{fig:cliques}(d) effectively enforce spatial coherence of the recovered support.  When such an overlapping group-sparsity model is used, all pixels in a clique tend to be either zero or non-zero at the same time (see, e.g., \cite{bach2011convex}).  Since each pixel shares multiple overlapping cliques with its neighbors,  this regularizer suppresses ``rogue'' (or isolated) pixels from entering the support without their neighbors and hence, promotes smooth (contiguous) support boundaries.

\subsection{Applications}

The proposed regularizer \eqref{reg} can be used as a building block for various applications in computer vision, image processing, and compressive sensing. In what follows, we will focus on the following three imaging applications:

%\begin{itemize}[leftmargin=*]

\textit{1) Compressive sensing signal recovery:} Consider a signal $\mathbf{x}\in\mathbb{R}^N$ that is $K$-sparse, i.e., only $K \ll N$ entries of~$\vecx$ are non-zero. In the CS literature, the signal is acquired via $M < N$ linear projections $\mathbf{y} = \mathbf{\Phi} \mathbf{x}$. The $K$-sparse signal $\mathbf{x}$ can then be recovered if, for example, the matrix $\mathbf{\Phi}$ satisfies the \textit{2K-RIP} or similar conditions~\cite{baraniuk2010model,candes2006robust}. The underlying recovery problem is usually formulated as follows: 
\aln{
  \vecx\opt =&  \argmin_{\mathbf{x}\in \mathbb{R}^N}   \|\mathbf{y} -\mathbf{\Phi} \mathbf{x}\|_2^2 \,\, \st \!\! \|\mathbf{x}\|_0  = K. \label{csrec}
 }
% If the signal $K=\supp(\vecx)$ is so large that \matPhi is does not satisfy the 2K-RIP, then the signal may be unrecoverable. 
When the sparse signals are images, simple sparse recovery may not exploit the entire image structure; this is particularly true for background-subtracted surveillance video.  Background subtraction is used in applications where one is interested only in inferring foreground objects and activities.  Background subtraction is easily achieved in the compressive domain by computing the difference between adjacent image data or by subtracting a long term signal mean (or median). Background-subtracted frames are generally more sparse than frames containing background information, and can thus be reconstructed from far fewer measurements $M$. 

We propose to extend the problem in (\ref{csrec}) by adding a regularizer of the form \eqref{reg} to promote correlation in the support set of the foreground objects. The optimization problem defined in (\ref{csrec}) is non-convex and is commonly solved using greedy algorithms  \cite{tropp2007signal, needell2009cosamp, cevher2009sparse}. We will show that the use of our prior~\eqref{reg} leads to faster signal recovery with a small number of measurements compared to existing methods.

\textit{2) Total-variation denoising:} Total variation (TV) denoising restores a noisy image $\vecy$ (e.g., vectorized image) by finding an image that lies close to $\vecy$ in an \elltwo-norm sense, while simultaneously having small total variation; this can be accomplished by solving 
   \aln{ 
   \vecx\opt = \argmin_{\vecx\in\reals^N} \textstyle \half \|\vecx-\vecy\|^2 + \lambda\|\nabla_d \vecx\|_1, \label{rof}
   }
 where $\nabla_d: \reals^N \to \reals^{2N}$ is a discrete gradient operator that acts on an $N$-pixel image, and produces a stacked horizontal and vertical gradient vector containing all first-order differences between adjacent pixels.  TV-based image processing assumes that images have a piecewise constant representation, i.e., the gradient is sparse and locally contiguous~\cite{rudin1992nonlinear,goldstein2009split}.
% \begin{align}
%  \argmin_{\mathbf{z}\in \mathbb{R}^N, \mathbf{x}\in \mathbb{R}^{2N}} D(\mathbf{y},\mathbf{\Phi x},\mathbf{z}) &=  ||\mathbf{y}-\mathbf{z}||_2^2 + \lambda \sum_{i} ||\tilde{\mathbf{x}}_{i}||_2 \nonumber \\
%  s.t. \quad \mathbf{z} &=\mathbf{\Phi} \mathbf{x}, \quad \tilde{\mathbf{x}}_{i} = [x_i, x_{i+N}] \label{eq3}
% \end{align}
%  \begin{align}
%  \argmin_{\mathbf{z}\in \mathbb{R}^N, \mathbf{x}\in \mathbb{R}^{2N}} D(\mathbf{y},\mathbf{\Phi x},\mathbf{z}) &=  ||\mathbf{y}-\mathbf{z}||_2^2 + \lambda \sum_{i} ||\tilde{\mathbf{x}}_{i}||_2 \nonumber \\
%  s.t. \quad \mathbf{z} &=\mathbf{\Phi} \mathbf{x}, \quad \tilde{\mathbf{x}}_{i} = [x_i, x_{i+N}] \label{eq3}
% \end{align}
%From (\ref{eq3}) we note that the total variation prior is indeed $\ell_1/\ell_2$ norm prior over gradients of individual pixel. 
%
Numerous generalizations of TV exist, including the recently proposed vectorial TV for color images \cite{bresson2008fast,ono2014decorrelated}. Such regularizers are of the form of \eqref{rof} merely by changing the definition of the discrete gradient operator.

We propose to extend total variation by penalizing the gradient of cliques in order to enforce a greater degree of spatial coherence.  In particular, we consider 
\aln{\vecx\opt = \argmin_{\vecx\in\reals^N} \textstyle  \half \|\vecx-\vecy\|^2 + J(\nabla_d \vecx), \label{rof2} }
   where  $J(\cdot)$ denotes the regularizer \eqref{reg}.  Furthermore, we explore formulations where the discrete gradient operator is given by the decorrelated color TV operator described in \cite{ono2014decorrelated}. With our approach, we also show the application of proposed structured sparsity prior on 3-D blocks. Note that \cite{selesnick2013total,liu2015image} explores the use of \mbox{1-D} and \mbox{2-D} overlapping group sparsity for TV image denoising, but using a majorization-minimization algorithm combined with ADMM.
 
\textit{3) Robust PCA (RPCA):} Suppose $\mathbf{Y} = [\mathbf{y}_1,  \dots, \mathbf{y}_L]$ is a matrix of $L$ measurement vectors, and $\mathbf{Y}$ is the sum of a low rank matrix $\mathbf{Z}$ and a sparse matrix $\mathbf{X}.$ For this case,  Cand\`es \etal show that exact recovery of these components is possible using the following  formulation \cite{candes2011robust}:
\begin{align} \begin{split}
  \{\hat\bZ,\hat\bX\} =  & \argmin_{\mathbf{Z},\mathbf{X}\in \mathbb{R}^{N \times L}}  \|\mathbf{Z}\|_* + \lambda \|\mathbf{X}\|_1  \\
  & \st  \!\mathbf{Y} = \mathbf{Z} + \mathbf{X}. \label{eq4}
  \end{split}
 \end{align}
The nuclear-norm in \eqref{eq4} promotes a low rank solution for \matZ; the \ellone-norm penalty on  promotes sparsity in \matX.  For this reason the solution to  \eqref{eq4} is sometimes referred to as a sparse-plus-low-rank decomposition.
A well-known application of RPCA is background subtraction in videos with a stationary background.  For such datasets, the shared background in the frames $\{y_i\}$ can be represented using a low-rank subspace.  The moving foreground objects often have sparse support, and thus are absorbed into the sparse term \matX.

We propose to replace the \ellone{}-norm regularization prior on~$\bX$ in~(\ref{eq4}) with the proposed regularizer in \eqref{reg}; this enables us to promote spatial smoothness in the support set of the foreground objects.
%
%, as observed in \cite{cevher2008compressive,zhou2009face}. 
%
%While the original formulation of RPCA showed state-of-the-art performance on some background subtraction tasks \cite{candes2011robust}, simple sparse models ignore the 
%
%To preserve this spatial correlation, 
%
Here, we build on the work of \cite{gao2012block}, where structured sparsity with non-overlapping blocks is used in RPCA for foreground detection, and \cite{yao2014foreground}, where a hybrid of ALM and network flow methods \cite{mairal2010network} are used to solve $\ell_1/\ell_\infty$ regularized RPCA problems.
%\end{itemize}

\subsection{Optimization Algorithms}
We now develop efficient numerical methods for solving problems involving the regularizer \eqref{reg}.  
%The optimal numerical approach to support-set regularization clearly depends on the structure of the data term \eqref{general}. In general, no single method can handle every possible choice efficiently.  For this reason, we propose various different numerical approaches and discuss when to use each of them. 
A common approach to enforce group sparsity in the statistics literature is consensus ADMM \cite{deng2013group,boyd2011distributed}, which we will briefly discuss in Section \ref{sec:stuff}.  For image processing and vision applications, where the datasets as well as the cliques tend to be large, the high memory requirements of ADMM render this approach unattractive.  As a consequence, we propose an alternative method that uses fast convolution algorithms to perform gradient descent that exhibits low memory requirements and requires low computational complexity.  In particular, our approach is capable of handling large-scale problems, such as those in video applications, which are out of the scope of memory-hungry ADMM algorithms.

We note that numerical methods for overlapping group sparsity have been studied in the context of statistical regression \cite{jacob2009group,yuan2011efficient,boyd2011distributed}, but for different purposes. Yuan \etal~\cite{yuan2011efficient} solves the regression variable selection problem using an accelerated gradient descent approach, whereas Deng \cite{deng2013group} and Boyd \cite{boyd2011distributed} use consensus ADMM, which does not scale to high-dimensional problems. Compared to these methods, our approach provides significant speedups (see Section~\ref{sec:experiments}). 

\subsubsection{Proximal Minimization and ADMM}
\label{sec:stuff}
The simplest instance of the problem \eqref{general} is the proximal operator for the penalty term $J$ in \eqref{reg}, defined as follows:
\begin{equation}\label{prox} 
\prox_{J}(\vecv,\lambda)  = \argmin_\vecx \|\vecx-\vecv\|^2 + \lambda J(\vecx) .
%   &= \argmin_\vecx \|\vecx-\vecv\|^2 + \lambda \sum_{c \in \mathcal{C}} ||\mathbf{x}_c||_2.
\end{equation}
Proximal minimization is a key sub-step in a large number of  numerical methods.  For example, the ADMM for TV  minimization \cite{rudin1992nonlinear,goldstein2009split} requires the computation of the proximal operator of the \ellone-norm.  For such methods, the regularizer~\eqref{reg} is easily incorporated into the numerical procedure by replacing this proximal minimization with \eqref{prox}.

  In the simplest case where the cliques in $\mathcal{C}$ are small and no other regularizers are needed, the proximal minimization~\eqref{prox} can be computed using ADMM \cite{boyd2011distributed,goldstein2009split}.   Similar approaches have been used for other applications of overlapping group sparsity \cite{deng2013group}.
It is key to realize that the regularization term in (\ref{prox}) can be reformulated as follows: 
\begin{align}
 \hat\vecx= \argmin_{\mathbf{x}\in \mathbb{R}^N} \, \|\mathbf{x}-\mathbf{v}\|_2^2 &  + \lambda \sum_{i=1}^{s}\sum_{c\in \mathcal{C}_i} \|\mathbf{x}_c\|_2. \label{eq7}
\end{align}
Here, $\mathcal{C}_1, \dots, \mathcal{C}_s$ are clique subsets for which the cliques in $\mathcal{C}_i$ are {\em disjoint}. For example, consider the case where the set of cliques contains all $2\times 2$ image patches as shown in Figure \ref{fig:cliques}(d). For such a scenario, we need four subsets of disjoint cliques to represent every possible patch.  The reformulated problem for the example graph will be of the form \eqref{eq7} with $s=4.$  In general, if cliques are formed by translating an $l\times l$ patch, $l^2$ subsets of cliques are required so that every subset contains only disjoint cliques.
%The reformulated problem for the example graph will be,
%\begin{align}
% \argmin_{\mathbf{x}\in \mathbb{R}^N} \quad ||\mathbf{x}-\mathbf{v}||_2^2 &+ \lambda \left(\sum_{c\in \mathcal{C}_1} ||\mathbf{x}_c||_2 + \sum_{c\in \mathcal{C}_2} ||\mathbf{x}_c||_2  \right. \nonumber \\
% &+ \left. \sum_{c\in \mathcal{C}_3} ||\mathbf{x}_c||_2 + \sum_{c\in \mathcal{C}_4} ||\mathbf{x}_c||_2\right) \label{eq7a}
%\end{align}

To apply ADMM to this problem, we need to introduce~$s$ auxiliary variables $\mathbf{z}^1, \dots, \mathbf{z}^s$ each representing a copy of the original pixel values. 
%The consistency in solution between the auxiliary variables is linked through the original variable~$\mathbf{x}$ using equality constraints. 
%
The resulting problem is
\begin{align} %
\begin{split}
\{\hat\vecx,\hat\vecz^i\,\forall i\} = & \argmin_{\mathbf{x}, \{\mathbf{z}^i\}_{i=1}^{s} }  \|\mathbf{x}-\mathbf{v}\|_2^2 + \lambda \sum_{i=1}^{s}\sum_{c\in \mathcal{C}_i} \|\mathbf{z}^i_c\|_2  \\
  &\!\! \st \mathbf{z}^i = \mathbf{x}, \,\, \forall i. \label{constrained}
  \end{split}
\end{align}
This is an example of a consensus optimization problem, which can be solved using ADMM (see \cite{deng2013group} for more details).   
%
%The resulting algorithm is listed in Algorithm \ref{alg:admm}.
%
%The \textit{scaled} augmented Lagrangian for \eqref{constrained} is given by \cite{boyd2011distributed} 
%\begin{align}
% \mathcal{L}(\vecx, \vecz, \vecb) = \|\mathbf{x}-\mathbf{v}\|_2^2 &+  \sum_{i=1}^s \sum_{c\in \mathcal{C}_i} \lambda \|\mathbf{z}^i_c\|_2 \nonumber \\
% &+ \sum_{i=1}^s \frac{\rho}{2} \|\mathbf{z}^i - \mathbf{x} + \vecb^i\|^2_2, \label{lag}
%\end{align}
%where $\rho$ is a scalar penalty parameter and the vectors $\{\vecb^i\}$ denote Lagrange multipliers (also known as Bregman vectors) for the linear constraints.   It can be shown that a saddle point of the Lagrangian \eqref{lag} corresponds to a solution of \eqref{constrained}.  Such a saddle point is found using the ADMM, which alternates between minimizing \eqref{lag} for \vecx and \vecz, and then updating \vecb.  This procedure is listed in Algorithm \ref{alg:admm}.
%
%
An important property of this ADMM reformulation is that each vector $\vecz^i$ can be updated in closed form---an immediate result of the disjoint clique decomposition.  

\subsubsection{Forward-Backward Splitting (FBS) with Fast Fourier Transforms}

The above discussed ADMM approach has several drawbacks.  First, it is difficult to incorporate more regularizers (in addition to the support regularizer $J$) without the introduction of an excessive amount of additional auxiliary variables.  Furthermore, the method becomes inefficient and memory intensive for large clique sizes and large data-sets (as it is the case for multiple images). For instance in RPCA, if the cliques are generated by $l\times l$ patches, $l^2$ variables $\{\vecz^i\}$ are required, each having the same dimensionality as original image data-set $NL$. Additionally, the dual variables for each equality constraints in \eqref{constrained} will require another $l^2NL$ storage entries. 
As a consequence, for large values of $l,$ the memory requirements of ADMM become prohibitive.

We propose a new forward-backward splitting algorithm that exploits fast convolution operators and  prevents the excessive memory overhead of ADMM-based methods. To this end, we propose to ``smoothen'' the objective via \emph{hyperbolic regularization} of the \elltwo-norm as 
\begin{align} \label{smooth}
 \|\mathbf{x}_c\|_2 \approx \|\mathbf{x}_c\|_{2,\epsilon} = \sqrt{x_1^2+\dots+x_n^2 + \epsilon^2}
\end{align}
for some small $\epsilon>0.$  For the sake of clarity, we describe the forward-backward splitting approach in the specific case of robust PCA. Note, however, that other regularizers are possible with only minor modifications. 

Using the proposed support prior  (\ref{eq4}), we write
\begin{align}
\{\hat\bZ,\!\hat\bX\}\!=\!\argmin_{\mathbf{Z},\mathbf{X}} \|\mathbf{Z}\|_* \!\! \textstyle+\! \lambda J_\epsilon(\bX) \!+\! \frac{\mu}{2} \|\mathbf{Y}\!-\!\mathbf{Z}\!-\!\mathbf{X}\|_F^2  \label{general2}
\end{align}
where 
\begin{equation} \textstyle J_\epsilon(\bX) = \sum_{t=1}^L \sum_{c\in \mathcal{C}} \|\mathbf{X}_{t,c}\|_{2,\epsilon} \label{smoothed}\end{equation}
is the smoothed support regularizer, and $\mathbf{X}_{t,c}$ refers to the clique $c$ drawn from column $t$ of $\mathbf{X}.$ We note that this formulation differs from that in Liu \etal \cite{liu2013robust}, where the structured sparsity is induced across columns of $\mathbf{X}$ rather than blocks, and is solved using conventional ADMM.

The forward-backward splitting (or proximal gradient) method is a general framework for minimizing  objective functions with two terms \cite{DBLP:journals/corr/GoldsteinSB14}.   For the problem \eqref{general2}, the method alternates between gradient descent steps that only act on the smooth terms in \eqref{general2}, and a backward/proximal step that only acts on the nuclear norm term.  The gradient of the (smoothed) proximal regularizer in  \eqref{general2} is given column-wise (i.e., image-wise) by
 \begin{equation}\textstyle  \nabla J_\epsilon(\matX_t) = \sum_{\substack{c\in C  }} \matX_{t,c}  \|\matX_{t,c}\|_{2,\epsilon}^{-1}. \label{grad}\end{equation}
 The gradient formula \eqref{grad} requires the computation of the sum \eqref{smoothed} for every clique $c,$ and then, a summation over the reciprocals of these sums; this is potentially expensive if done in a na\"ive way.  Fortunately, every block sum can be computed simultaneously by squaring all of the entries in $\matX,$ and then convolving the result with a block filter.  The result of this convolution contains the value of $\|\matX_{t,c}\|_{2,\epsilon}^2$ for all cliques $c.$  Each entry in the result is then raised to the $-\nicefrac{1}{2}$ power, and convolved again with a block filter to compute the entries in the gradient \eqref{grad}.   Both of these two convolution operations can be computed quickly using fast Fourier transforms (FFTs), so that the computational complexity  becomes independent of clique size.
 
Algorithm \ref{alg:fbs} shows the pseudocode for solving (\ref{general2}).  In Steps 1 and 2, the values of \matX and \matZ are updated using gradient descent on  \eqref{general2}, ignoring the nuclear norm regularizer.  Step 3 accounts for the nuclear-norm term using its proximal mapping, which is given by
  $$\prox_*(\bQ, \delta) =  \matU (\text{sign}(\matS)\circ\max\{|\matS|-\delta,0\})\matV^T,$$
  where $\bQ =\matU\matS\matV^T$ is a singular value decomposition of~$\bQ,$ $|\matS|$ denotes element-wise absolute value, and $\circ$ denotes element-wise multiplication.
  \begin{algorithm}[t]
 \caption{Forward-backward proximal minimization }
 \label{alg:fbs}
  \textbf{Input:} $\mathbf{Y}, \mu > 0, \lambda, \mathcal{C}_i , \alpha>0$\\
  \textbf{Initialize:} $\mathbf{X}^{(0)} = \mathbf{0}$, $\mathbf{Z}^{(0)} = \mathbf{0}$\\
  \textbf{Output:} $\mathbf{X}^{(n)}, \mathbf{Z}^{(n)} $ 
  \begin{algorithmic}[1]
  \WHILE {not converged}
    \STATE \textbf{Step 1:} Forward gradient descent on $X$,
    \STATE $\matX^{(n)}_k =$ \parbox[t]{.7\linewidth}{
    $\matX^{(n-1)}_k - \alpha\lambda  \nabla J_\epsilon(\matX)$  \\$+ \alpha\mu(\matY_k-\matZ^{(n-1)}_k-\matX^{(n-1)}_k)$}
    \STATE \textbf{Step 2:} Forward gradient descent on $\matZ$,
    \STATE $\,\,\, {\matZ}^{(n)}_k = \matZ^{(n-1)}_k + \alpha \mu(\matY_k-\matZ^{(n-1)}_k-\matX^{(n-1)}_k)$
    \STATE \textbf{Step 3:} Backward gradient descent on $\matZ$,
    \STATE $\quad \mathbf{\matZ}^{(n)} = \prox_*(\matZ^{(n)}, \alpha) $
  \ENDWHILE
  \end{algorithmic}
\end{algorithm}

%<<<<<<< HEAD
The forward-backward splitting (FBS) procedure in Algorithm~\ref{alg:fbs} is known to converge for sufficiently small stepsizes $\alpha$ \cite{beck2009fast}.  Practical implementations of FBS \ref{alg:fbs} include adaptive stepsize selection \cite{wright2009sparse}, backtracking line search, or acceleration~\cite{beck2009fast}.  We use the  FASTA solver from  \cite{DBLP:journals/corr/GoldsteinSB14}, which  combines such acceleration techniques.
%=======
%The forward-backward splitting (FBS) procedure in \ref{alg:fbs} is known to converge for sufficiently small stepsizes $\alpha$ \cite{beck2009fast}.  Practical implementations of FBS Algorithm \ref{alg:fbs} include adaptive stepsize selection \cite{wright2009sparse}, backtracking line search, or acceleration~\cite{beck2009fast}.  We use the  FASTA solver from  \cite{DBLP:journals/corr/GoldsteinSB14}, which  combines such acceleration techniques.
%>>>>>>> 9587c67bea5995fcc23821c85059d3b548bfa655

We note that FBS \ref{alg:fbs} only requires a total of $4NL$ storage entries for $\mathbf{X}, \mathbf{Y}, \mathbf{Z}$ and gradient $\nabla J_\epsilon(\mathbf{X})$. However, in order to solve RPCA formulation using ADMM we require $2l^2NL$ storage entries for auxiliary variables (as discussed before) and $4NL$ storage entries for the variables $\mathbf{X}, \mathbf{Y}, \mathbf{Z}$ and dual variable of $\mathbf{Y} = \mathbf{X} + \mathbf{Z}$, leading to total of $(2l^2 + 4)NL$ storage entries. Since the memory usage and runtime of FBS is independent of the clique size, the advantage of FBS over ADMM is much greater for larger cliques.

\subsubsection{Matching Pursuit Algorithm}

For compressive-sensing problems involving large random matrices, {\em matching pursuit} algorithms (such as CoSaMP~\cite{needell2009cosamp}) are an important class of sparse recovery methods.  When signals have structured support, model-based matching pursuit routines have been proposed that require non-convex minimizations over Markov random fields~\cite{cevher2009sparse}.  In this section, we propose a model-based matching pursuit algorithm that achieves structured compressive signal recovery using {\em convex} sub-steps for which global minimizers are efficiently computable. 

The proposed method, Convex Lattice Matching Pursuit (CoLaMP), is a greedy algorithm that attempts to solve 
\begin{equation}
\begin{split}
\hat\vecx= & \argmin_\vecx \|\matPhi \vecx - \vecy\|^2_2 + \lambda J(\vecx)\\
&  \!\!\!\st \!\! \|\vecx\|_0\le K. \label{colampeq}
\end{split}
\end{equation}
%  $$\minimize\!\!\!\!\! \|\matPhi \vecx - \vecy\|^2_2 + \lambda J(\vecx) \st \!\! \|\vecx\|_0\le K  \label{s0}.$$ 
%<<<<<<< HEAD
 The complete method is listed in Algorithm \ref{alg:colamp}. In Step 1, CoLaMP proceeds like other matching pursuit algorithms; the unknown signal is estimated by multiplying the residual by the adjoint of the measurement operator.  In Step 2, this estimate is refined by solving a support regularized problem of the form \eqref{prox}.  We solve this problem either via ADMM  or the FBS method in Algorithm \ref{alg:fbs}). In Step 3, a least-squares (LS) problem is solved to identify the signal that best matches the observed data, assuming the correct support was  identified in Step 2.  This LS problem is solved by a conjugate gradient method. Finally, in Step 4, the residual (the discrepancy between $\matPhi \vecx$ and the data vector \vecy) is calculated.  The algorithm is terminated if the residual becomes sufficiently small or a maximum number of iterations is reached.
%=======
% The complete method is listed in Algorithm \ref{alg:colamp}. In Step 1, CoLaMP proceeds like other matching pursuit algorithms; the unknown signal is estimated by multiplying the residual by the adjoint of the measurement operator.  In Step 2, this estimate is refined by solving a support regularized problem of the form \eqref{prox}.  This problem is solved by either the ADMM routine or the FBS method in Algorithm \ref{alg:fbs})  In Step 3, a least-squares problem is solved in order to identify the signal that best matches the observed data, assuming the correct support was  identified in Step 2.  This least-squares problem is efficiently solved by a conjugate gradient method. Finally, in Step 4, the residual (the discrepancy between $\matPhi \vecx$ and the data vector \vecy) is calculated.  The algorithm is terminated as soon as the residual becomes sufficiently small or a maximum number of iterations is reached.
%>>>>>>> 9587c67bea5995fcc23821c85059d3b548bfa655

\begin{algorithm}[t]
 \caption{CoLaMP - Convex Lattice Matching Pursuit}
 \label{alg:colamp}
  \textbf{Input:} $\mathbf{y}, \mathbf{\Phi}, K, \lambda, \epsilon$\\
  \textbf{Initialize:} $\mathbf{x}^{(0)} = \mathbf{0}$, $\mathbf{s}^{(0)} = \mathbf{0}, \mathbf{r}^{(0)} = \mathbf{y}$ \\
  \textbf{Output:} $\mathbf{x}^{(n)}$
  \begin{algorithmic}[1]
  \WHILE {$\, n \le$ {\em max\_iterations} and $\|\mathbf{r}^{(n)}\|_2 > \epsilon$} 
  \STATE \textbf{Step 1:} Form temporary target signal
    \STATE $\quad \mathbf{v}^{(n)} \gets \Phi^T \mathbf{r}^{(n-1)} + \mathbf{x}^{(n-1)}$
  \STATE \textbf{Step 2:} Refine signal support using convex prior\\
  \STATE  $\quad \vecx_r^{(n)} = \argmin_\vecx \|\vecx-\mathbf{v}^{(n)}\|_2^2+\lambda J(x),$
  \STATE  $\quad \vecs \gets  \supp(\mathbf{x}_r^{(n)}) $
  \STATE \textbf{Step 3:} Estimate target signal
   \STATE $\quad$ Solve $ \matPhi_s^T \matPhi_s \vecx_s  =  \matPhi_s^T \vecy,$ with $\matPhi_s = \matPhi(:,\vecs)$
   \STATE $\quad$ Set all but largest $K$ entries in $ \vecx_s$ to zero,
    \STATE $\quad \vecx^{(n)}(s)=  \vecx_s(s) $
  \STATE \textbf{Step 4:} Calculate data residual
    \STATE $\quad \mathbf{r}^{(n)} \gets \mathbf{y} - \mathbf{\Phi} \mathbf{x}^{(n)}$    
  \STATE $\quad n \gets n+1$
  \ENDWHILE
  \end{algorithmic}
\end{algorithm}

CoLaMP has several desirable properties. First, the support set regularization (Step 2) helps to prevent signal support from growing quickly, and thus minimizes the cost of the least-squares problem in Step 3.  Secondly, the use of a convex prior guarantees that a global minimum is obtained for every subproblem in Step 2, regardless of the considered clique structure. This is in stark contrast to other model-based recovery algorithms, such as LaMP\footnote{It is possible to restrict LaMP to planar Ising models, in which case a global optimum is computable \cite{cevher2009sparse}.}, and  model-based CoSaMP \cite{baraniuk2010model}, which requires the solution to non-convex optimization problems to enforce structured support set models.

%------------------------------------------------------------------------
%\begin{figure*}[tbp]
%\begin{center}
%   \includegraphics[width=0.8\linewidth,height=2in]{qualitative_bs.pdf}
%   \vspace{-.25cm}
%\end{center}
%   \caption{\vspace{0.2cm}Compressed sensing recovery results for background subtracted images using $M=3K$.}
%\label{fig:qualitative_bs}
%\end{figure*}

\begin{figure*}[tbp]
\begin{center}
   \includegraphics[width=\linewidth,height=2in]{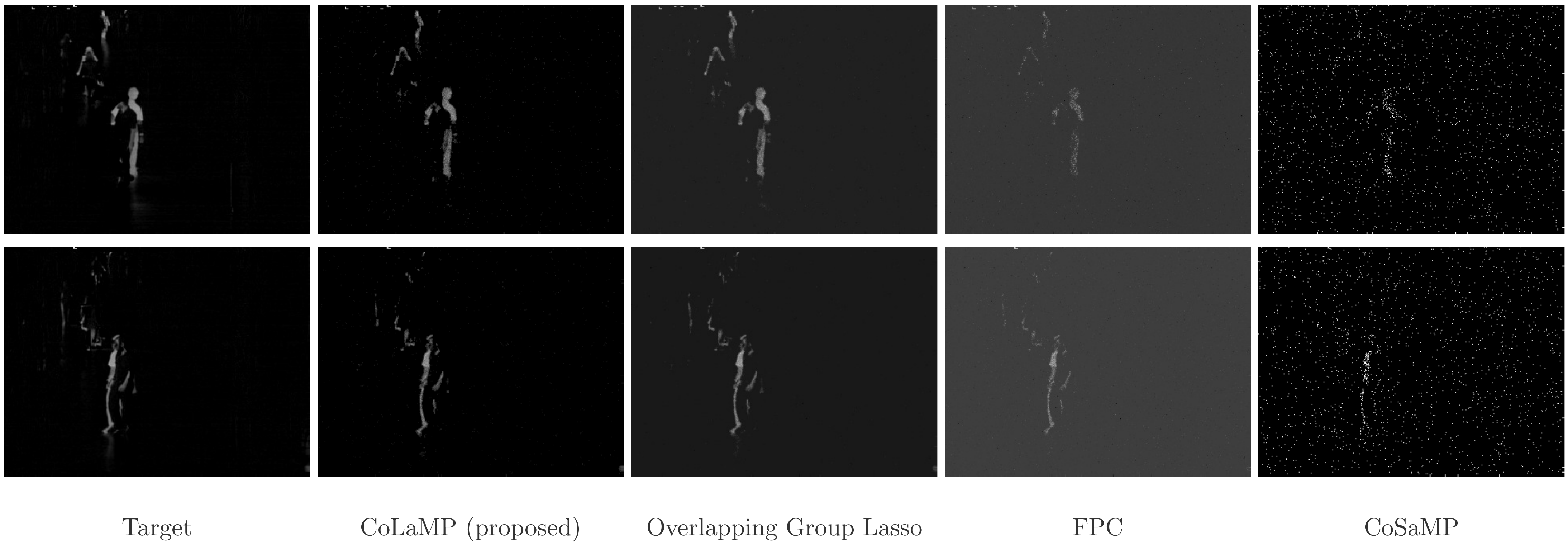}
   \vspace{-.25cm}
\end{center}
   \caption{\vspace{0.2cm}Compressed sensing recovery results for background subtracted images using $M=3K$.}
\label{fig:qualitative_bs}
\end{figure*}

\section{Numerical Experiments}
\label{sec:experiments}

We now apply the proposed regularizer to a range of datasets  to demonstrate its efficacy for various  applications. Unless stated otherwise, we showcase our algorithms using overlapping cliques of size $2\times2$ as shown in Figure~\ref{fig:cliques}(b). Note that the numerical algorithms need not be restricted to those discussed above as different schemes (such as primal-dual decomposition) are needed for different situations.

\subsection{Compressive Image Recovery}

%<<<<<<< HEAD
We first consider the recovery of background-subtracted images from compressive measurements. We use the ``walking2" surveillance video data \cite{WuLimYang13} with frames of dimension $288\times384$. Test data is generated by choosing two frames from a video sequence and computing the pixel-wise difference between their intensities. We compare the output of our proposed CoLaMP algorithm to that of other state-of-the-art recovery algorithms, such as overlapping group lasso \cite{leigsnering2014multipath}, fixed-point continuation (FPC) \cite{hale2007fixed} and CoSaMP \cite{needell2009cosamp}. Note that CoSaMP defines the support set using the $2K$ largest components of the error signal. The group lasso algorithm is equivalent to minimizing the objective in \eqref{colampeq} using variational method. Unlike the CoLaMP algorithm, this method does not consider prescribed signal sparsity $K$. An example recovery using $M = 3K$ measurements is shown in Figure~\ref{fig:qualitative_bs}. The sparsity level $K$ is chosen such that the recovered images account for $97\%$ of the compressive signal energy. The average $K$ across datasets is $2800$ and we fix $\lambda = 2.$ Note that the spatially clustered pixels are recovered almost perfectly. Further, we randomly generated 50 such test images from the above dataset and compared the performance of the CoLaMP, group lasso, and FPC algorithms under varying numbers of measurements from $1K$ to $5K$. The performance is measured in terms of the magnitude of reconstruction error normalized by the original image magnitude.  Results are shown in Figure \ref{fig:comparison} (left). We clearly see that 
%for small numbers of measurements, 
the proposed smooth sparsity prior significantly improves the reconstruction quality over FPC. Furthermore, our algorithm is $7\times$ faster than the group lasso algorithm. For $M/K = 3$, the average runtime is $215s$ for CoLaMP and $1510s$ for the group lasso algorithm.
%=======
%We first consider the recovery of background-subtracted images from compressive measurements. We use the well-known airport surveillance video data \cite{li2004statistical} with frames of dimension $144\times176$. Test data is generated by choosing two frames from a video sequence and computing the pixel-wise difference between their intensities. We compare the output of our proposed CoLaMP algorithm to that of other state-of-the-art recovery algorithms, such as fixed-point continuation (FPC) \cite{hale2007fixed} and CoSaMP \cite{needell2009cosamp}. Note that CoSaMP defines the support set using 2K largest components of the error signal. An example recovery using $M = 3K$ measurements is shown in Figure \ref{fig:qualitative_bs}. The sparsity level $K$ is chosen such that the recovered images account for $97\%$ of the compressive signal energy and we fix $\lambda = 7.$ Note that the spatially clustered pixels are recovered almost perfectly. Further, we randomly generated 50 such test images from the above dataset and compared the performance of the CoLaMP and FPC algorithms under varying numbers of measurements from $1K$ to $5K$. The performance is measured as magnitude of reconstruction error normalized by the original image magnitude.  The associated results are shown in Figure \ref{fig:comparison} (left). We clearly see that for small numbers of measurements, the proposed smooth sparsity prior significantly improves the reconstruction quality over FPC.
%>>>>>>> 9587c67bea5995fcc23821c85059d3b548bfa655

\subsection{Robust Signal Recovery}

\begin{figure*}[!tbp]
\begin{center}
\vspace{-3mm}
   \includegraphics[width=0.8\linewidth,height=2in]{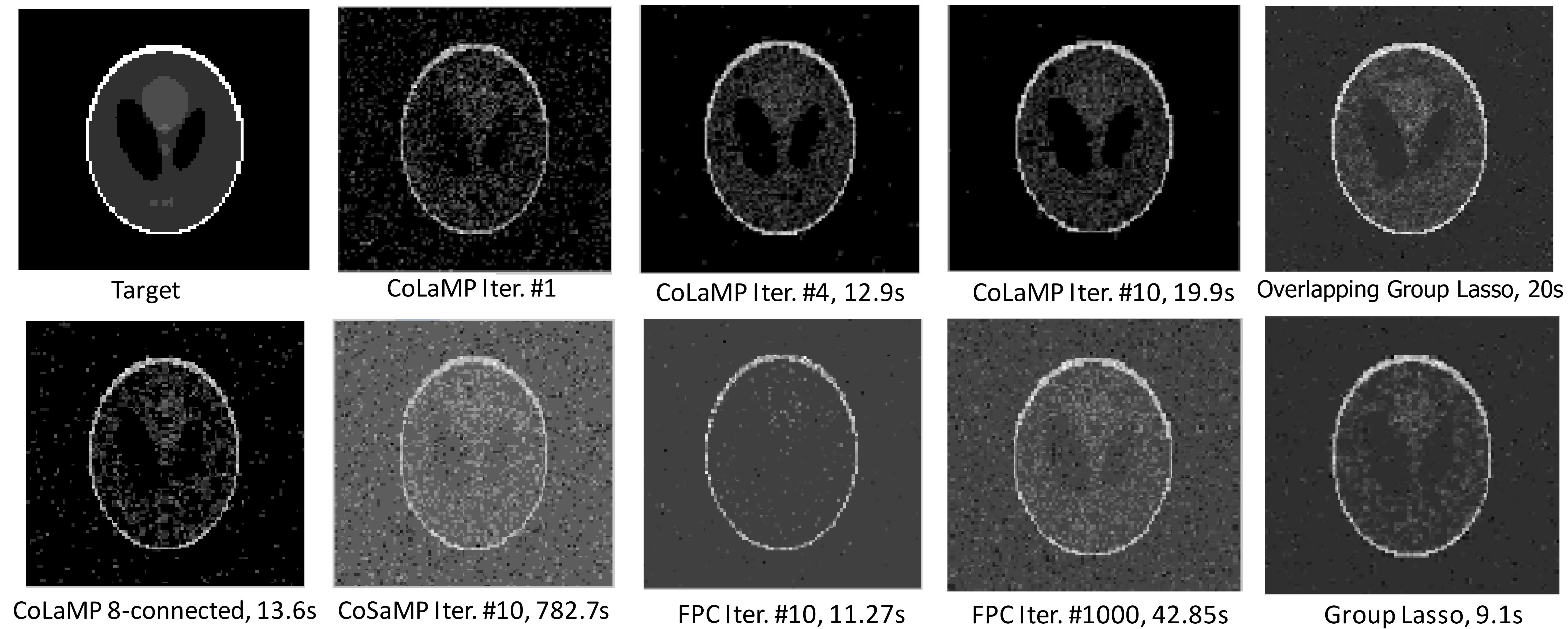}
   \vspace{-.25cm}
\end{center}
   \caption{Robust recovery results for the phantom image from a noisy compressed signal.}
\label{fig:qualitative_cs}
\end{figure*}

\begin{figure*}[!tbp]
\vspace{-2mm}
\centering
%\begin{subfigure}{.33\textwidth}
  \centering
  \includegraphics[width=.33\linewidth]{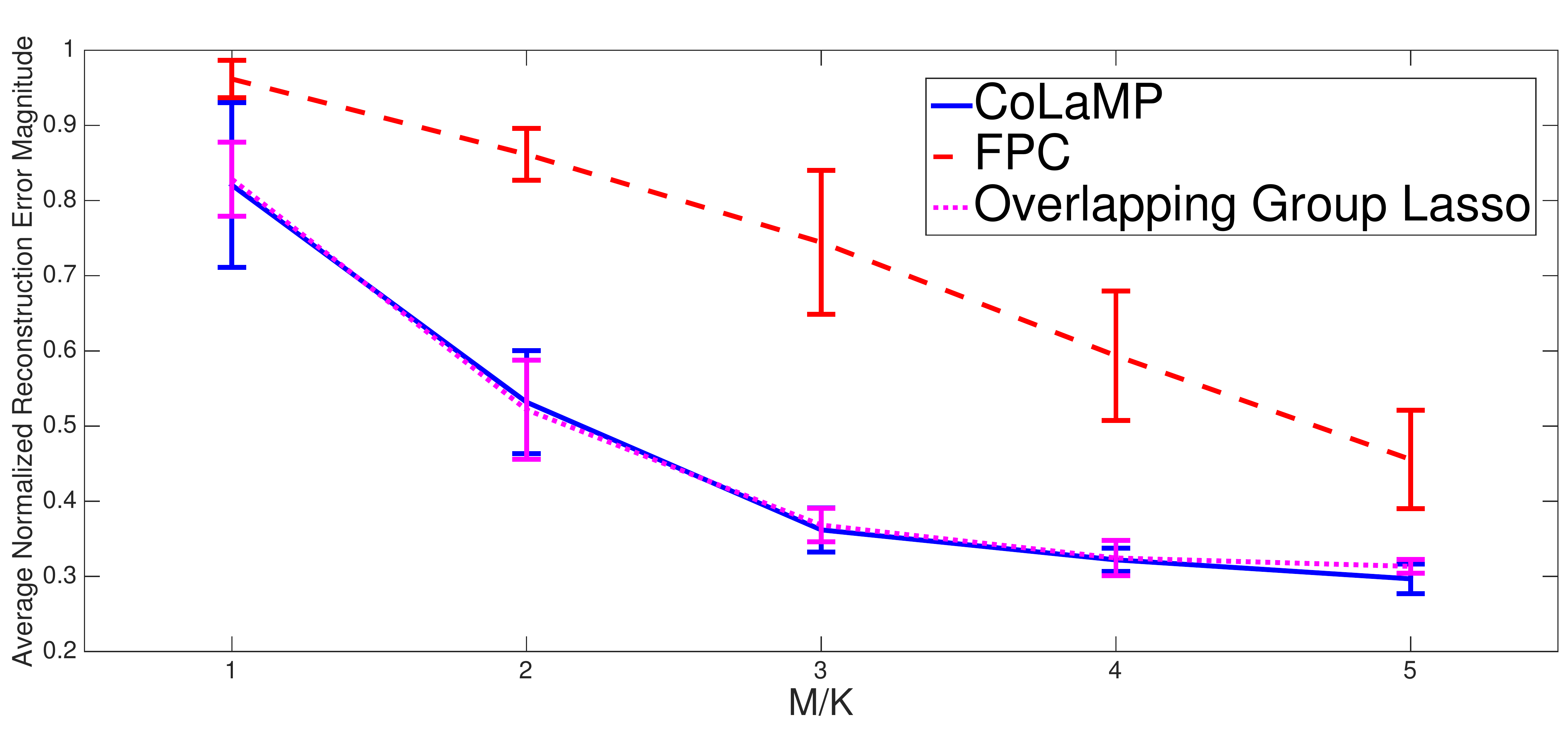}
 % \caption{}
%  \label{fig:quantitative_bs}
%\end{subfigure}%
%\begin{subfigure}{.33\textwidth}
%  \centering
  \includegraphics[width=.33\linewidth]{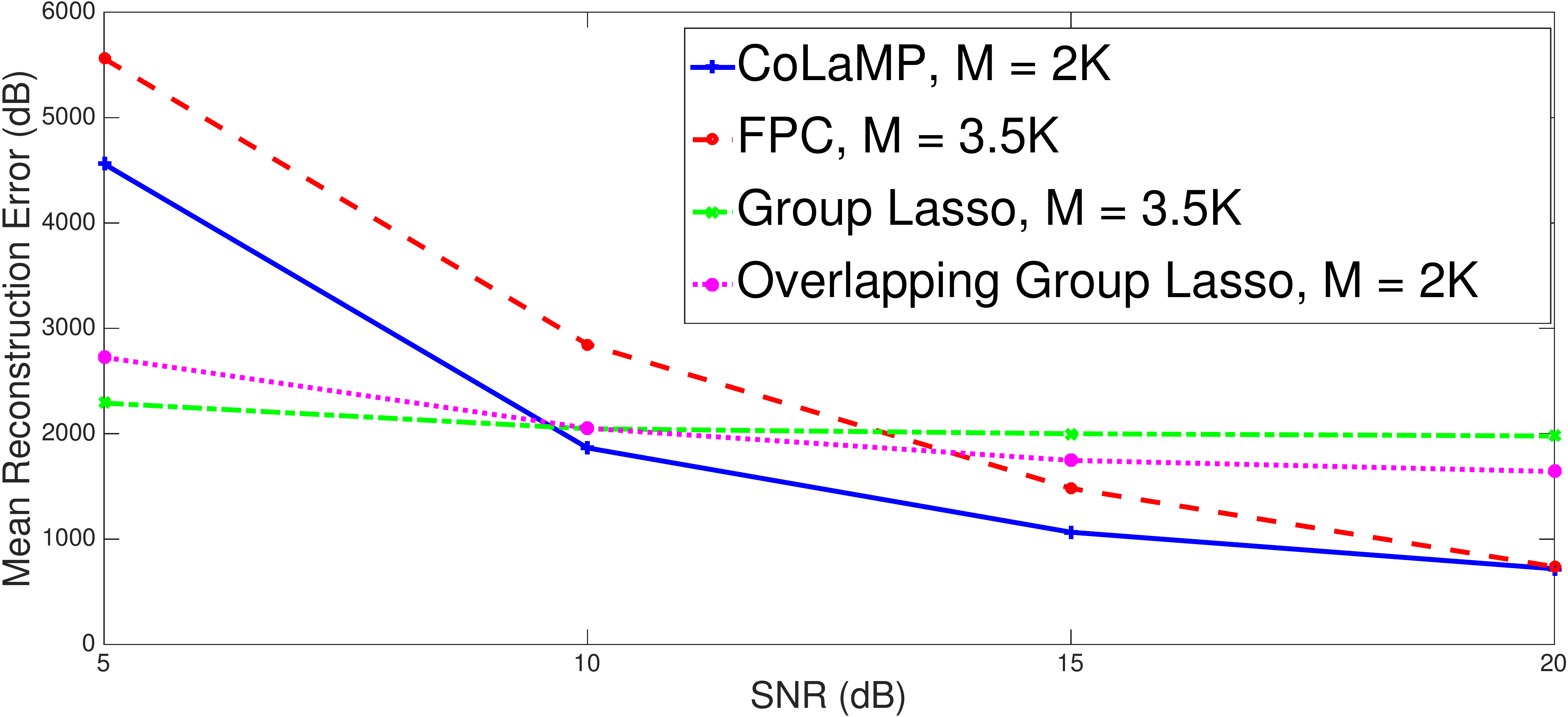}
 % \caption{}
%  \label{fig:quantitative_cs}
%\end{subfigure}
%\begin{subfigure}{.33\textwidth}
 % \centering
   \includegraphics[width=.33\linewidth]{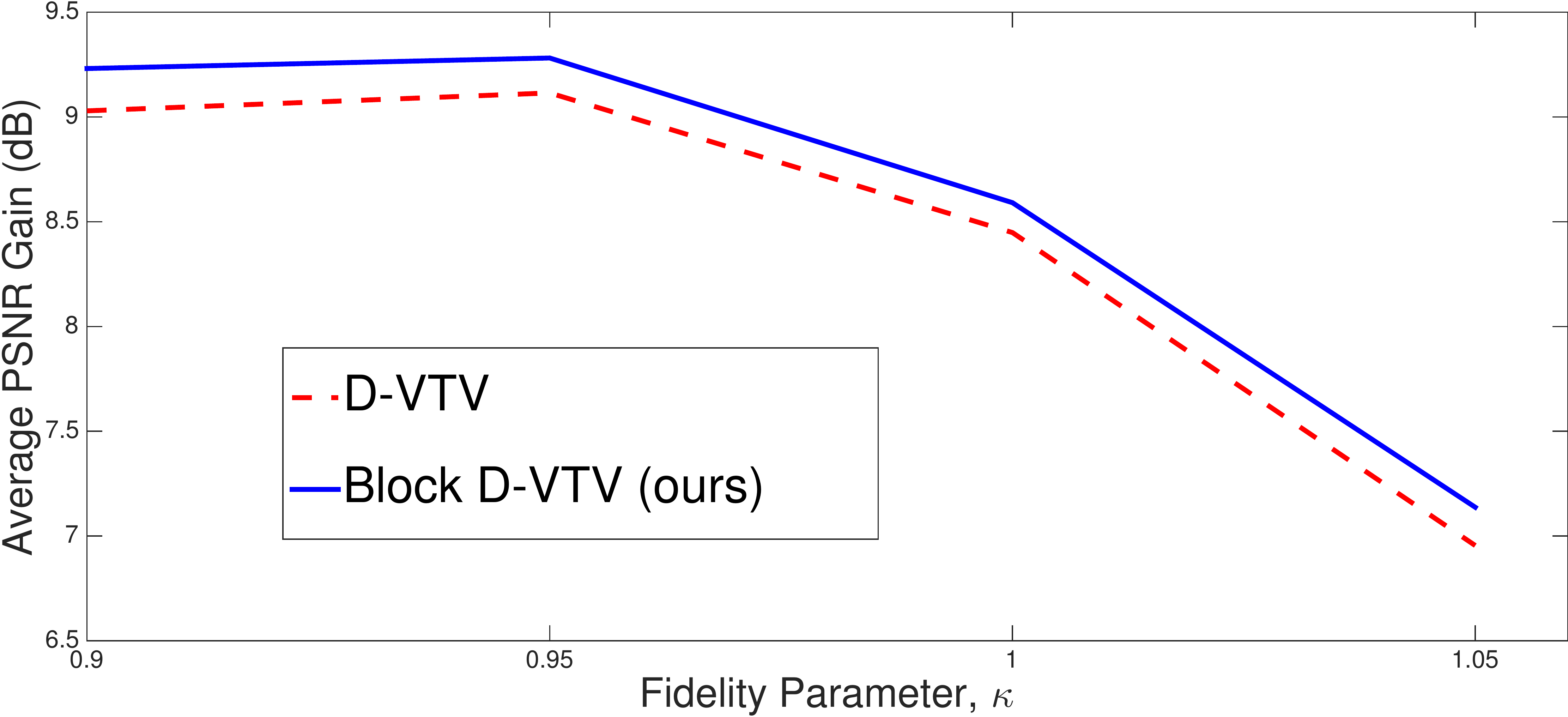}
 % \caption{}
%  \label{fig:quantitative_dn}
%\end{subfigure}%
\caption{Quantitative Comparison: (left) Recovery performance of compressed sensing on background subtracted images; (center) Robust compressed sensing recovery error at various SNR; (right) Average denoising gain in PSNR (dB) for various values of $\kappa$}
\label{fig:comparison}
\end{figure*}

We next showcase the suitability of CoLaMP for signal recovery from noisy compressive measurements. We consider a $100\times100$ Shepp\textemdash Logan phantom image with a support size of $K = 2636.$ A Gaussian random measurement matrix was used to sample $M=2K$ measurements, and the measurements were corrupted with additive white Gaussian noise. The signal-to-noise ratio of the resulting measurements is $10$\,dB. Figure \ref{fig:qualitative_cs} shows the original and recovered images for various recovery algorithms. We also show the output from the first few iterates of the CoLaMP algorithm. The support of the target signal is almost exactly recovered within four iterations of CoSaMP and stabilizes by the end of 10 iterations. Figure \ref{fig:qualitative_cs} also shows the recovery times of various algorithms running on the same laptop computer. CoLaMP is approximately 40$\times$ faster than the CoSaMP algorithm and it is at least 2$\times$ faster than FPC.

%On the same laptop computer, running time of CoLaMP for 10 iterations outperforms the CoSaMP-10 and FPC-1000 methods as shown in Figure \ref{fig:qualitative_cs} \cs{what exactly is shown in this figure? it's not clear; mention that the recovery times are shown and say how much faster we are}.

To enable a fair comparison, we also show the output obtained with CoLaMP using the 8-connected pixel clique in Figure \ref{fig:cliques}(c), as well as the output of the group lasso algorithm \cite{yuan2006model}, where each clique is of size $2\times2$. All these algorithms and our proposed method are implemented using ADMM. Not surprisingly, while all these algorithms beat CoLaMP in terms of runtime, their recovered signals do not match CoLaMP in terms of perceived closeness to target signal as shown in Figure \ref{fig:qualitative_cs}. The CoLaMP results are  regularized by $\lambda_0=16.$ We then used an increasing value of $\lambda_n=1.02^n\lambda_0$ where $n$ is the iteration number. In practice, we obtain better results if $\lambda$ increases over time as it will heavily penalize sparse, blocky noise. For all other algorithms, we used the implementations provided by the authors.

For detailed quantitative comparisons, we repeat the above experiment using 100 Gaussian random measurement matrices and record the average reconstruction error with SNR varying from $5$\,dB to $20$\,dB. 
%For reference, $10$\,dB SNR of measured signal corresponds to a noise amplitude of approximately $16.$ 
For each algorithm, $M$ is fixed to the minimal measurement number required to give close to perfect recovery in the presence of noise. For CoLaMP and overlapping group lasso, we set $M=2K,$  whereas for FPC and non-overlapping group lasso we set $M=3.5K.$ Figure \ref{fig:comparison} (center) illustrates that CoSaMP outperforms FPC at all SNRs even with $1.5K$ fewer measurements. Group lasso performs best at low SNR while its performance flattens out starting at $10$\,dB.

\vspace{-2mm}
\subsection{Color Image Denoising}
We now consider a variant of the denoising problem (\ref{rof2}) where the image gradient is defined over color images using the decorrelated vectorized TV (D-VTV) proposed in  \cite{ono2014decorrelated}
\begin{align}
\hat\vecx = & \argmin_{\mathbf{x}\in \mathbb{R}^{3N}} \sum_{c\in\mathcal{C}} \lambda\|\nabla_d 
\vecx^{\ell}_c\|_2 + \|\nabla_d \vecx^{ch}_c\|_2 \nonumber \\
  & \!\! \st \|\mathbf{x}-\mathbf{y}\|_2 \leq \kappa m. \label{general3}
 \end{align}
Here, $\nabla_d\vecx^{\ell}\in\mathbb{R}^{2N}$ and $\nabla_d\vecx^{ch}\in\mathbb{R}^{4N}$ represent the stacked gradients of luminance and chrominance channels of the input color image, the constant $m$ depends on the noise level, and $\kappa$ is a fidelity parameter. To solve this problem numerically, we use the primal-dual algorithm described in \cite{ono2014decorrelated}, but we replace the shrinkage operator with the proximal operator \eqref{prox} to adapt our clique-based regularizer.  
  
\begin{figure*}[!tbp]
\begin{center}
   \includegraphics[width=1.0\linewidth,height=2.4in]{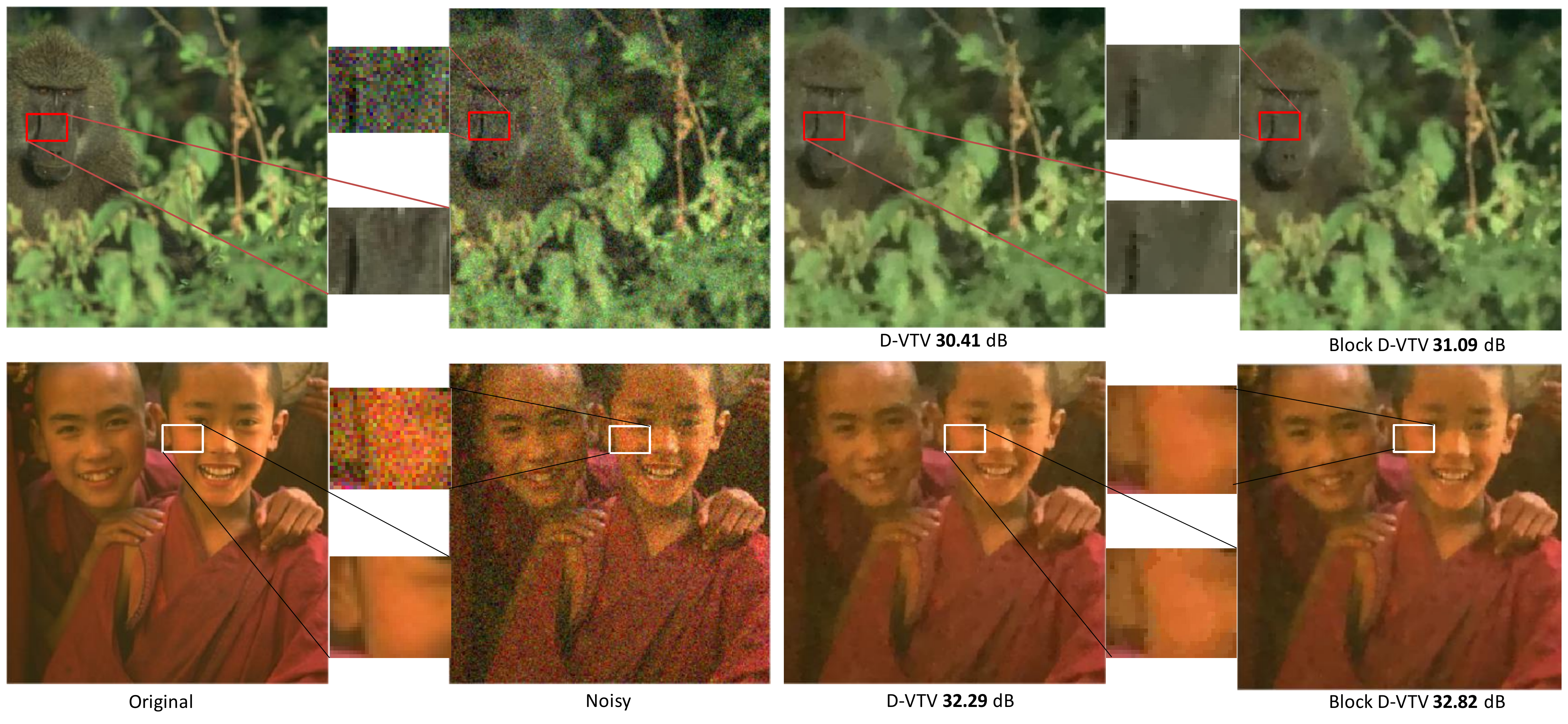}
   \vspace{-.6cm}
\end{center}
   \caption{Restoration of noisy images using Block D-VTV and existing D-VTV (best viewed in color).}
\label{fig:qualitative_dn}
\end{figure*}

  Following a protocol similar to D-VTV \cite{ono2014decorrelated}, we conduct experiments using 300 images from the Berkeley Segmentation Database \cite{MartinFTM01}. Noisy images with average PSNR 20\,dB are obtained by adding white Gaussian noise. The resulting denoised output of our method (Block D-VTV)  is compared to D-VTV in Figure \ref{fig:qualitative_dn}. The zoomed-in version reveals that our method exhibits less uneven color artifacts and less pronounced staircasing artifacts than the D-VTV results. A quantitative comparison measured using average PSNR gain (in dB) is drawn in Figure \ref{fig:comparison} (right) for various values of $\kappa$. Our method outperforms D-VTV by $0.25$\,dB. Also note that our method, Block D-VTV, obtains relatively better PSNR gain than the state-of-the-art D-VTV method at smaller values of $\kappa$. This observed gain is significant because smaller $\kappa$ values lead to a tighter fidelity constraint and thus a smaller solution space around the noisy input. In such situations, Block D-VTV helps to improve image quality by leveraging input from neighboring pixels.

\subsection{Video Decomposition}

\begin{figure*}[t]%[!tbp]
\begin{center}
   \includegraphics[width=1.0\linewidth,height=2.7in]{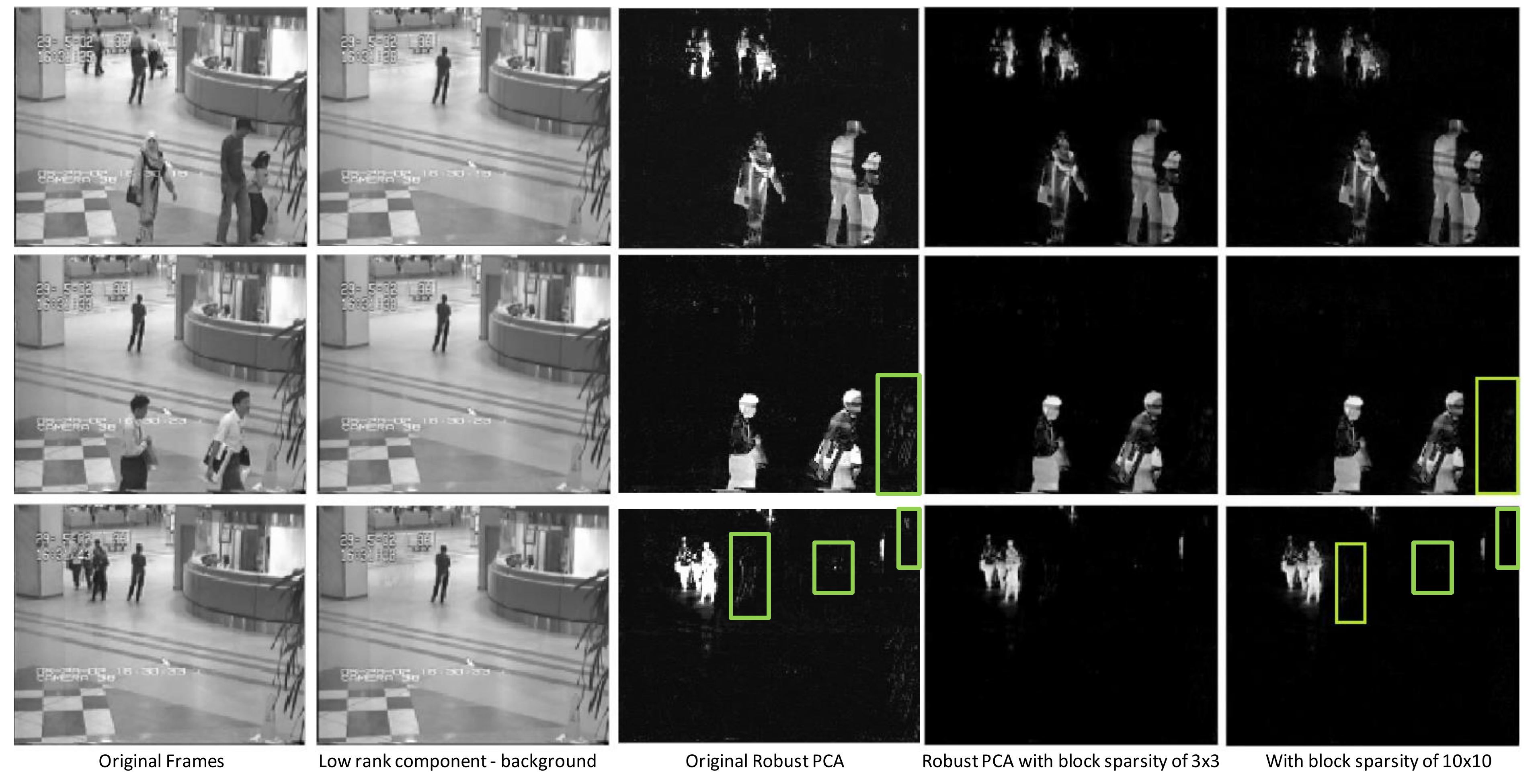}
   \vspace{-.6cm}
\end{center}
   \caption{\vspace{1mm}Sparse-and-low-rank decomposition using original robust PCA and proposed approach.}
\label{fig:qualitative_rpca}
\end{figure*}

We finally consider the robust PCA (RPCA) problem for structured sparsity of size $10\times 10$ as formulated in (\ref{general2}) and using Algorithm 2. We consider the same airport surveillance video data \cite{li2004statistical} as in \cite{candes2011robust} with frames of dimension $144\times176$. For a clique formed from $l\times l$ patches, we observed that $\lambda = 1/(l\sqrt{n_1})$ works best for our experiments as opposed to $\lambda = 1/\sqrt{n_1}$ used in \cite{candes2011robust}. This is because each element of the matrix $\mathbf{X}$ is shared by $l^2$ sparsity inducing terms. The resulting low rank components (background) and foreground  components of three such example video frames are shown in Figure \ref{fig:qualitative_rpca}. For all the approaches, the low rank components are nearly identical.  We observe that the rank of the low-rank component remains the same. As highlighted with the green box, the noisy sparse edges appearing in the original RPCA  disappear from the foreground component using our proposed method.  We also display the foreground component obtained using smaller overlapping cliques of size $3\times 3$, but solved using ADMM as opposed to forward-backward splitting (Algorithm \ref{alg:fbs}). 
%While running on the same laptop computer, we found that for clique size of $10\times 10$, the ADMM solver is 4x faster than forward-backward splitting but requires 50x more memory.
%
We found that for clique size of $10\times 10$ the ADMM method becomes intractable because it requires approximately 50$\times$ more memory than the proposed forward-backward splitting method with fast convolutions (i.e., $204NL$ vs.\ $4NL$).

%------------------------------------------------------------------------
%\vspace{-2mm}
\section{Conclusions}
\vspace{-2mm}
We have proposed a novel structured support regularizer for convex sparse recovery. Our regularizer can be applied to a variety of problems, including sparse-and-low-rank decomposition and denoising.  For compressive signal recovery using large unstructured matrices, our convex regularizer can be used to improve the recovery quality of existing matching-pursuit algorithms.  Compared to existing algorithms for this task, our proposed approach enjoys the capability of fast signal reconstruction from fewer measurements while exhibiting superior robustness against spurious artifacts and noise. For color image denoising, the restored images reveal more homogeneous color effects. For robust PCA, we achieve improved foreground-background separation with far fewer artifacts.
We envision many more applications that could benefit of the proposed regularizer, including deblurring and inpainting. More sophisticated directions include using support regularization for structured dictionary learning \cite{zhang2013learning} and multitask classification. 
%Here, the training label prior can help imposing structure on the sparse dictionary coefficients which may in turn influence the dictionary learning process. This is intended to improve the classification accuracy.

\section*{Acknowledgments}

The work of S.~Shah and T.~Goldstein was supported in part by the US National Science Foundation (NSF) under grant CCF-1535902 and by the US Office of Naval Research under grant N00014-15-1-2676. The work of C.~Studer was supported in part by Xilinx Inc., and by the US NSF under grants ECCS-1408006 and CCF-1535897.

%------------------------------------------------------------------------
%\clearpage
{\small
\bibliographystyle{ieee}
\bibliography{paperbib}
}

\end{document}